\documentclass[11pt]{article}

\usepackage[final]{acl}

\usepackage{times}
\usepackage{latexsym}

\usepackage[T1]{fontenc}
\usepackage[utf8]{inputenc}

\usepackage{microtype}

\usepackage{inconsolata}
\usepackage{float}
\usepackage[ruled,vlined]{algorithm2e}
\usepackage{microtype}
\usepackage{booktabs}
\usepackage{multirow}
\usepackage{graphicx}
\usepackage{placeins}
\usepackage{booktabs}

\usepackage[table]{xcolor}
\usepackage{threeparttable}
\usepackage{graphicx}
\usepackage{amsmath}
\usepackage{amssymb}
\definecolor{modelrow}{HTML}{EFEAF8}
\definecolor{oursrow}{HTML}{F4DCE8}
\title{STaR-Quant: State-Time Consistent Post-Training Quantization for Diffusion Large Language Models}

\author{
  Xin Yan$^{1}$ \quad Aqiang Wang$^{1}$ \quad Zhenglin Wan$^{2}$ \quad  Xingrui Yu$^{3}$\thanks{Corresponding author. Email: \texttt{yu\_xingrui@a-star.edu.sg}.}\quad Ivor Tsang$^{3}$ \\
  $^{1}$School of Artificial Intelligence, Beijing Normal University, Beijing, China \\
  $^{2}$Department of Computer Science, National University of Singapore, Singapore \\
  $^{3}$Centre for Frontier AI Research, Agency for Science, Technology and Research (A*STAR), Singapore \\
}
\begin{document}
\maketitle
\begin{abstract}
Diffusion large language models (DLLMs) have recently emerged as a promising alternative to autoregressive LLMs by generating text through iterative masked denoising with bidirectional context. 
However, their large model sizes and iterative denoising process introduce substantial memory and computational overhead, motivating post-training quantization for efficient deployment. 
In this paper, we identify two key challenges for low-bit DLLM quantization: state-dependent activation disparity and temporal error accumulation. 
Masked and unmasked tokens exhibit different activation distributions within each denoising step, while quantization errors can accumulate across steps during iterative decoding. 
 To address these challenges, we propose \textbf{STaR-Quant}, a state-time consistent PTQ framework for DLLMs. 
STaR-Quant introduces \textbf{State-Guided Activation Transformation} (SGAT) to assign masked and unmasked tokens to different activation transformation spaces with a unified static weight-side transformation. 
It further introduces \textbf{Temporal Attention Compensation} (TAC) to correct the quantized attention representation via a lightweight block-diagonal affine mapping. Experiments on representative DLLMs demonstrate that STaR-Quant consistently improves low-bit weight-activation quantization over strong PTQ baselines, while delivering up to 1.69$\times$ speedup and 3.14$\times$ memory saving over FP16 deployment.
\end{abstract}

\section{Introduction}

Diffusion large language models (DLLMs) have recently attracted increasing attention as a new paradigm for text generation~\cite{nie2025large,ye2025dream}. 
Inspired by diffusion processes, DLLMs formulate language generation as an iterative masked denoising task: a sequence is first partially or fully masked, and the model progressively recovers masked tokens through a reverse denoising process. 
Different from autoregressive LLMs that predict tokens strictly from left to right~\cite{touvron2023llama,bai2023qwen,grattafiori2024llama3}, DLLMs leverage bidirectional context at each denoising step and can recover multiple tokens in parallel. 
This generation mechanism provides flexible token refinement and stronger control over output structure, showing promising performance in general language modeling, reasoning, and code generation tasks~\cite{nie2025large,ye2025dream}.

Despite these advantages, the efficient deployment of DLLMs remains a challenge. 
The denoising process usually requires multiple iterative steps to obtain high-quality outputs, and reducing the number of decoded tokens per step further increases the average computation cost. Meanwhile, recent DLLMs continue to scale to billions of parameters, bringing substantial memory consumption and inference overhead. 
These characteristics make compression essential for deploying DLLMs on resource-constrained devices.

Post-Training Quantization (PTQ), which quantizes
weights and activations into low-precision formats, effectively reduces memory usage and computational overhead, achieving notable success in LLMs ~\cite{frantar2022gptq,xiao2023smoothquant,lin2024awq,ashkboos2024quarot}. 
However, directly applying these PTQ methods to DLLMs often leads to substantial performance degradation, especially under aggressive low-bit weight-activation quantization~\cite{lin2026quantization}.
This performance drop reveals a clear mismatch between conventional PTQ assumptions and diffusion-style language inference, motivating a quantization framework tailored to the masked denoising process of DLLMs.

\begin{figure}[t]
    \centering
    \includegraphics[width=\columnwidth]{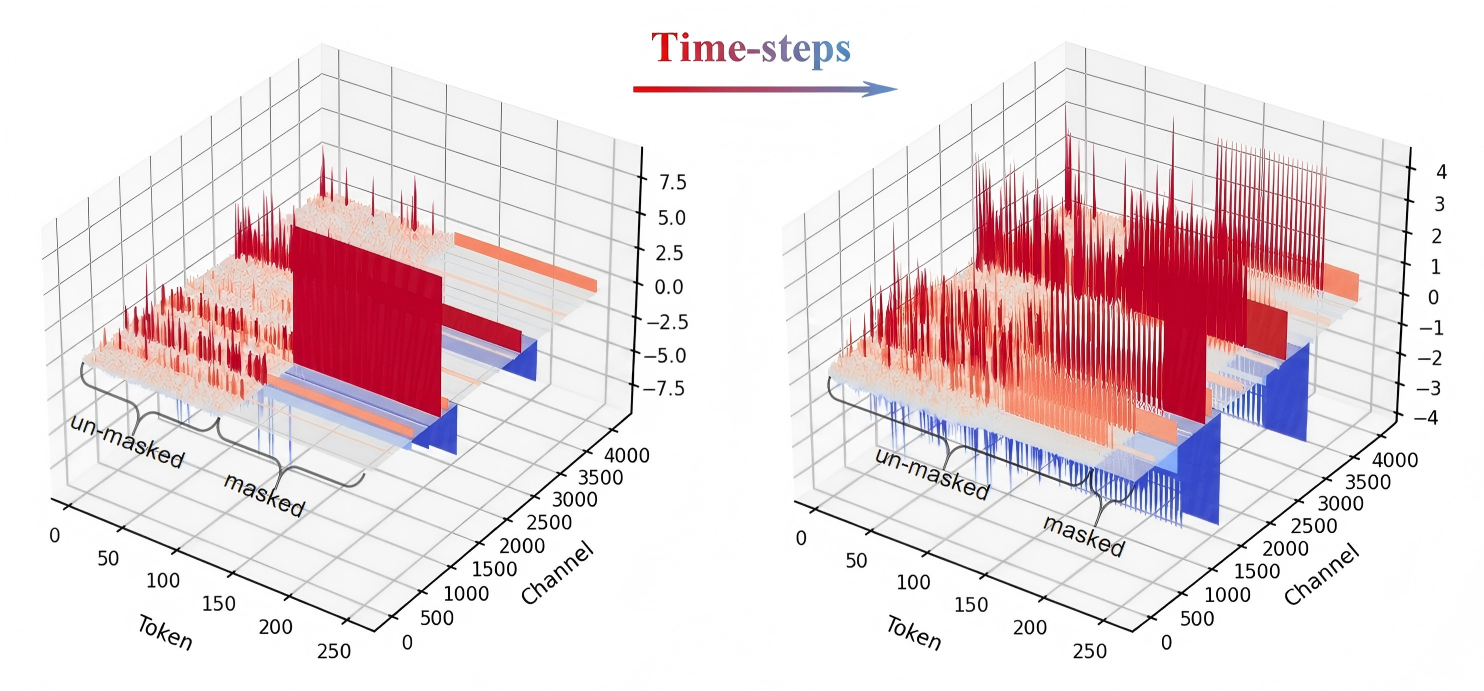}
    \caption{
    Activation distributions across token states and denoising steps in DLLMs. 
    Masked and unmasked tokens show distinct activation patterns, while the distributions also evolve over time, motivating state-time consistent quantization.
    }
    \label{fig:intro}
\end{figure}

In this work, we identify two key challenges for low-bit weight-activation quantization of DLLMs. 

The first challenge comes from \emph{state-dependent activation disparity}. 
During denoising, masked and unmasked tokens coexist in the same sequence. 
Masked tokens correspond to uncertain positions that need to be predicted, while unmasked tokens provide visible context. 
These two token states show different activation ranges and outlier patterns, as illustrated in Fig.~\ref{fig:intro}. 
As a result, using a single activation transformation or a unified quantization space for all tokens fails to smooth both states effectively.

The second challenge comes from \emph{temporal error accumulation}. 
In DLLMs, generation proceeds through iterative denoising, where the output of each step is reused in subsequent predictions. 
As a result, quantization errors are not confined to a single step, but can accumulate as the denoising process proceeds. 
Prior analysis shows that the matrix multiplication between the softmax attention map and value states is a critical error-prone operation under low-bit quantization~\cite{xu2025dllmquant}. 

To address these challenges, we propose \textbf{STaR-Quant}, a state-time consistent PTQ framework for DLLMs. 
For the state dimension, we introduce \textbf{State-Guided Activation Transformation (SGAT)}, which assigns masked and unmasked tokens to different activation-side transformation spaces. 
Specifically, the hidden dimension is decomposed into shared and state-specific subspaces, allowing each token state to be smoothed in a more suitable quantization space. Meanwhile, the weight side maintains a unified static transformation, avoiding duplicated transformed weights during inference. 
For the temporal dimension, we introduce \textbf{Temporal Attention Compensation} (TAC). Instead of changing the attention computation itself, TAC compensates its quantized output representation before it is fed into the attention output projection. Specifically, TAC applies a lightweight block-wise affine mapping to align the quantized attention representation with its full-precision counterpart, thereby reducing error accumulation along the denoising trajectory.

Our contributions are summarized as follows:
\begin{itemize}
    \item We identify two key challenges in low-bit DLLM quantization: state-dependent activation disparity between masked and unmasked tokens, and temporal error accumulation across iterative denoising steps.
    
    \item We propose State-Guided Activation Transformation, which maps masked and unmasked activations into shared and state-specific transformation spaces while keeping a unified static weight-side transformation.
    
    \item We introduce Temporal Attention Compensation, a closed-form block-wise affine compensation method that aligns quantized attention representations with full-precision counterparts before the attention output projection.
    
    \item Extensive experiments on representative DLLMs show that STaR-Quant improves low-bit weight-activation quantization performance over strong PTQ baselines.
\end{itemize}

\section{Related Work}
\subsection{Large Language Diffusion Models}
Large Language Diffusion Models have emerged as an alternative to autoregressive language modeling by generating text through iterative denoising rather than left-to-right token prediction~\cite{li2022diffusionlm,austin2021structured,lou2024discrete,sahoo2024simple}. 
Early efforts explored both continuous and discrete diffusion formulations for text generation, including Diffusion-LM for controllable generation~\cite{li2022diffusionlm} and D3PM for discrete state-space diffusion~\cite{austin2021structured}. 
Recent discrete diffusion language models further improved language modeling quality through score-entropy objectives~\cite{lou2024discrete} and masked discrete diffusion formulations ~\cite{shi2024simplified,sahoo2024simple}.
Building on these advances, large-scale diffusion LLMs such as DiffuLLaMA~\cite{gong2025scaling}, LLaDA~\cite{nie2025large}, and Dream~\cite{ye2025dream} scale diffusion-based generation to billion-parameter models. 
DiffuLLaMA adapts pretrained autoregressive LMs into diffusion models~\cite{gong2025scaling}, while LLaDA trains a diffusion LLM from scratch with a forward masking process and a reverse masked-token prediction objective~\cite{nie2025large}. 
Dream 7B further demonstrates competitive performance on general, mathematical, and coding tasks~\cite{ye2025dream}. 
Despite their parallel decoding and bidirectional context modeling advantages, DLLMs still face substantial deployment challenges due to large model size and high memory and computation 
cost.~\cite{gong2025scaling,nie2025large,ye2025dream}

\subsection{Post-Training Quantization for LLMs}

Post-training quantization (PTQ) has become a practical technique for compressing large language models, as it reduces memory footprint and inference cost without expensive retraining. 
Existing LLMs PTQ methods can be broadly divided into weight-only quantization and weight-activation quantization. 
Weight-only methods quantize model weights while keeping activations in high precision. 
GPTQ~\cite{frantar2022gptq} uses second-order information to compensate weight quantization errors, while AWQ~\cite{lin2024awq} preserves salient weights according to activation statistics. SmoothQuant~\cite{xiao2023smoothquant} shifts quantization difficulty from activations to weights through channel-wise scaling. 
Recent transformation-based methods improve low-bit quantization by smoothing activation distributions. QuaRot~\cite{ashkboos2024quarot} applies Hadamard rotations to obtain outlier-free representations, while DuQuant~\cite{2024duquant} combines outlier-aware rotation and channel permutation to better distribute activation outliers. Although these PTQ methods are effective for autoregressive LLMs, they do not explicitly consider the denoising process of DLLMs. 
Recent studies show that directly applying LLM-oriented PTQ methods to DLLMs still leads to clear degradation, especially under weight-activation quantization~\cite{lin2026quantization}. 
Different from autoregressive decoding, DLLMs contain masked and unmasked token states within each denoising step, and quantization errors may accumulate across iterative denoising steps. 

\section{Method}
\begin{figure*}[t]
    \centering
    \includegraphics[width=1.03\textwidth]{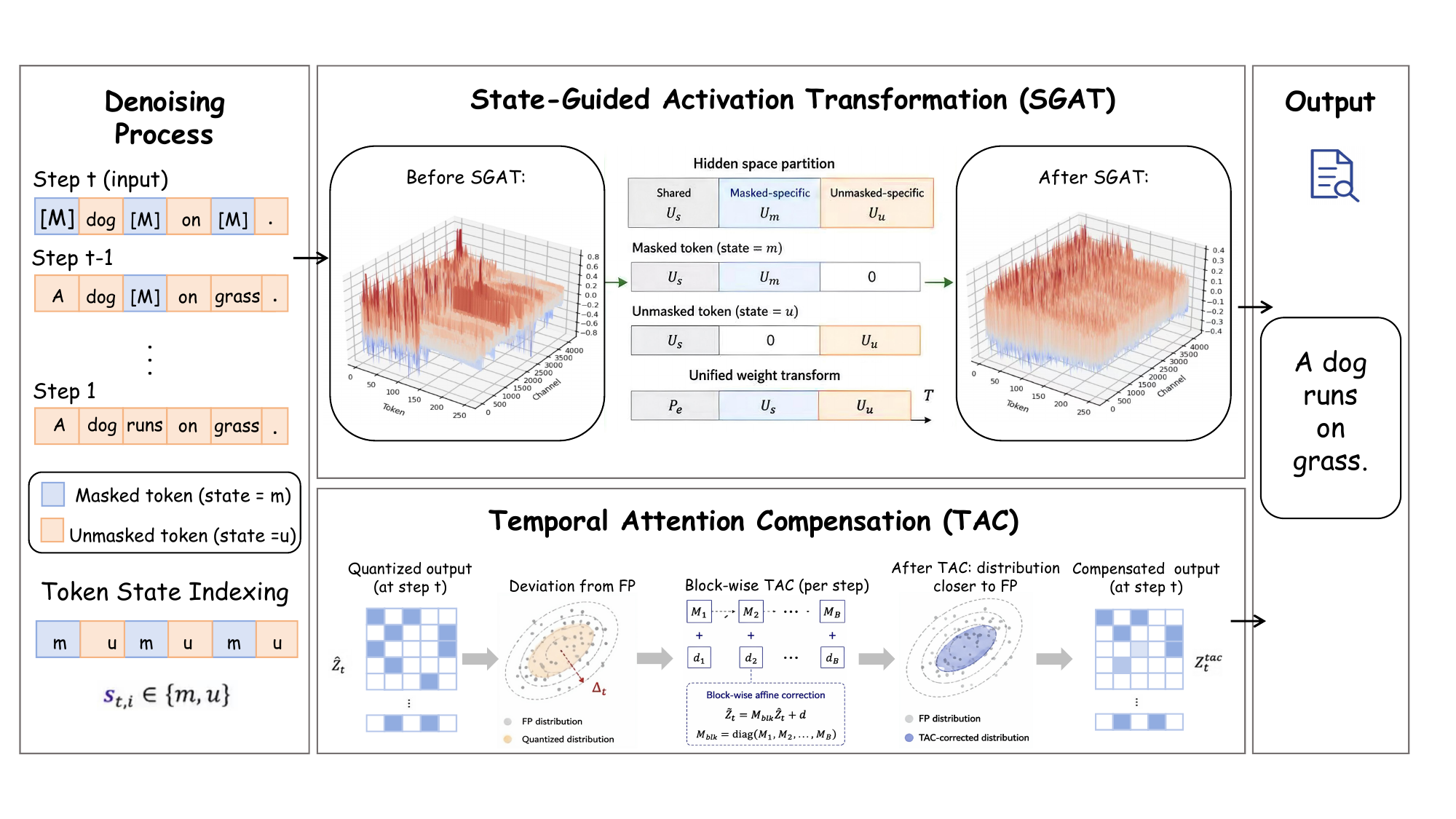}
    \caption{
    Overview of STaR-Quant. 
    Given a DLLM denoising sequence, tokens are first indexed by their mask states, i.e., masked and unmasked. 
    State-Guided Activation Transformation (SGAT) assigns the two token states to shared and state-specific activation subspaces while keeping a unified weight-side transformation. 
    Temporal Attention Compensation (TAC) further corrects the quantized attention representation before the output projection through block-wise affine compensation, reducing temporal quantization error along the denoising process.
    }
    \label{fig:overview}
\end{figure*}

\noindent\textbf{Overview.}
As shown in Fig.~\ref{fig:overview}, STaR-Quant addresses DLLM quantization from both the state and temporal dimensions. 
Given the intermediate sequence during denoising, we first index tokens according to their mask states. 
SGAT then transforms masked and unmasked activations into state-aware quantization spaces while sharing a unified transformed weight. 
After the attention computation, TAC compensates the quantized attention representation before the output projection, improving its consistency with the full-precision path. 
The two components together form a state-time consistent quantization framework for DLLMs.
\subsection{Preliminaries}
\label{sec:preliminaries}

We focus on post-training weight-activation quantization for DLLMs. 
For a linear layer with input activation $X\in\mathbb{R}^{L\times d}$ and weight $W\in\mathbb{R}^{C\times d}$, the full-precision computation is
\begin{equation}
Y = XW^\top .
\end{equation}
Given bit-width $b$, uniform quantization maps a floating-point tensor $X$ to a low-bit integer tensor:
\begin{equation}
X_{\mathrm{int}}
=
\mathrm{clamp}
\left(
\left\lfloor \frac{X}{s} \right\rceil + z,
0, 2^b-1
\right),
\end{equation}
where $s$ and $z$ are the quantization scale and zero point. 
The dequantized tensor is
\begin{equation}
\hat{X}=s(X_{\mathrm{int}}-z).
\end{equation}
For asymmetric quantization, $s$ and $z$ are computed as
\begin{equation}
s=\frac{X_{\max}-X_{\min}}{2^b-1},
\qquad
z=\left\lfloor -\frac{X_{\min}}{s}\right\rceil .
\end{equation}
Following common LLM PTQ settings, we use per-token quantization for activations and per-channel quantization for weights. 
We use $Q_a(\cdot)$ and $Q_w(\cdot)$ to denote activation and weight quantization, respectively. 
Then the quantized linear computation is approximated by
\begin{equation}
Y \approx Q_a(X)Q_w(W^\top).
\end{equation}

Low-bit activation quantization is sensitive to outliers and non-uniform value ranges. 
Transformation-based quantization mitigates this issue by mapping activations to a smoother quantization space. 
For a transformation matrix $P$, the linear computation can be rewritten as
\begin{equation}
Y
=
(XP)(P^{-1}W^\top)
\approx
Q_a(XP)Q_w(P^{-1}W^\top).
\end{equation}
This transformation smooths activations before quantization while preserving the original computation through the inverse transformation on the weight side.

\subsection{State-Guided Activation Transformation}
\label{sec:state_transform}

Transformation-based quantization typically couples each activation-side transform with a corresponding weight-side transform. 
For DLLMs, however, the same static weights process activations from different denoising states. 
Motivated by FreeAct~\cite{liu2026freeact}, we therefore make the activation transform state-conditioned: masked and unmasked tokens are routed to different transformed subspaces, while all tokens share one unified transformed weight.

For a quantized linear layer at denoising step $t$, let 
$X_t\in\mathbb{R}^{L\times d}$ denote the input activation. 
SGAT transforms activations into smoother quantization spaces before low-bit quantization, with the goal of reducing state-dependent outliers in DLLM activations. 
Since masked and unmasked tokens exhibit different activation patterns during denoising, we first split tokens according to their current states:
\begin{equation}
\mathcal{I}_s^t=\{i\mid s_{t,i}=s\}, \quad
X_t^s=X_t[\mathcal{I}_s^t], \quad
s\in\{m,u\},
\end{equation}
where $m$ and $u$ denote masked and unmasked tokens, respectively. 
The two activation groups are transformed and quantized separately, and their outputs are later scattered back to the original token order.

To smooth common and state-specific activation components separately, we construct the transformation basis with shared and state-specific subspaces:
\begin{equation}
d=d_{\mathrm{sh}}+d_m+d_u, \quad
P=[U_{\mathrm{sh}}, U_m, U_u].
\end{equation}
Here, $U_{\mathrm{sh}}\in\mathbb{R}^{d\times d_{\mathrm{sh}}}$ denotes the shared subspace basis, while 
$U_m\in\mathbb{R}^{d\times d_m}$ and 
$U_u\in\mathbb{R}^{d\times d_u}$ denote the masked-specific and unmasked-specific bases. 
The shared basis captures common activation components, whereas the state-specific bases absorb different outlier patterns from masked and unmasked tokens.

Following this decomposition, the state-conditioned transformations are implemented by binary subspace gates:
\begin{equation}
\begin{aligned}
G_m &= \mathrm{diag}(\mathbf{1}_{d_{\mathrm{sh}}}, \mathbf{1}_{d_m}, \mathbf{0}_{d_u}), \\
G_u &= \mathrm{diag}(\mathbf{1}_{d_{\mathrm{sh}}}, \mathbf{0}_{d_m}, \mathbf{1}_{d_u}).
\end{aligned}
\end{equation}
Equivalently, masked and unmasked tokens use
\begin{equation}
\begin{aligned}
P_m &= PG_m = [U_{\mathrm{sh}}, U_m, 0], \\
P_u &= PG_u = [U_{\mathrm{sh}}, 0, U_u].
\end{aligned}
\end{equation}
The zero-padded parts suppress the subspace assigned to the other state, avoiding interference between state-specific components. 
For state $s$, the activation-side transformation is
\begin{equation}
\Phi_s(X_t^s)=X_t^sPG_s,\quad s\in\{m,u\}.
\end{equation}
Thus, both states share the common basis $U_{\mathrm{sh}}$, while each state is smoothed in its own state-specific quantization subspace.

On the weight side, we keep a unified static transformation to match the fixed linear weights used across denoising steps. 
Let $P_e$ denote the weight-side transformation associated with the full basis $P$. 
For an orthogonal basis, $P_e=P^\top$; more generally, $P_e$ is the corresponding weight-side transform. 
For a linear layer with weight $W$, the transformed quantized weight is computed once:
\begin{equation}
\widetilde{W}=Q_w(P_eW^\top).
\end{equation}
The state-guided quantized projection is then
\begin{equation}
\hat{Y}_t^s
=
Q_a\!\left(\Phi_s(X_t^s)\right)\widetilde{W},
\quad s\in\{m,u\}.
\end{equation}
Finally, $\hat{Y}_t^m$ and $\hat{Y}_t^u$ are scattered back to their original token positions to form $\hat{Y}_t$. 
In this way, SGAT uses state-conditioned activation transformations to smooth activations, while keeping one shared transformed weight for all token states. The computation equivalence of this shared weight-side formulation is analyzed in Appendix~\ref{app:sgat_consistency}.
\paragraph{Optimization objective.}
Under the state-guided structure above, we optimize the transformation parameters by minimizing the layer-wise quantization error. 
For a denoising step $t$, let $Y_t^{\mathrm{FP}}$ denote the full-precision output of the linear layer, and let $\hat{Y}_t$ denote the output produced by the state-guided quantized projection. 
The objective is defined as
\begin{equation}
\mathcal{L}_{\mathrm{state}}
=
\mathbb{E}_{t}
\left\|
Y_t^{\mathrm{FP}}-\hat{Y}_t
\right\|_F^2 .
\end{equation}
Since SGAT computes the quantized projection separately for masked and unmasked tokens, the objective can be equivalently written as
\begin{equation}
\mathcal{L}_{\mathrm{state}}
=
\mathbb{E}_{t}
\sum_{s\in\{m,u\}}
\left\|
Y_t^{s,\mathrm{FP}}
-
Q_a\!\left(X_t^sPG_s\right)\widetilde{W}
\right\|_F^2 .
\end{equation}
This optimization learns the shared and state-specific subspaces under a unified transformed weight, reducing the state-wise quantization error introduced during DLLM denoising. 
After optimization, the weight-side transform is folded into the quantized weight, so inference only applies the state-conditioned activation transform followed by the quantized linear computation. More details are provided in Appendix~\ref{app:details}.
\subsection{Temporal Attention Compensation}
\label{sec:tac}

\begin{figure}[t]
    \centering
    \includegraphics[width=\columnwidth]{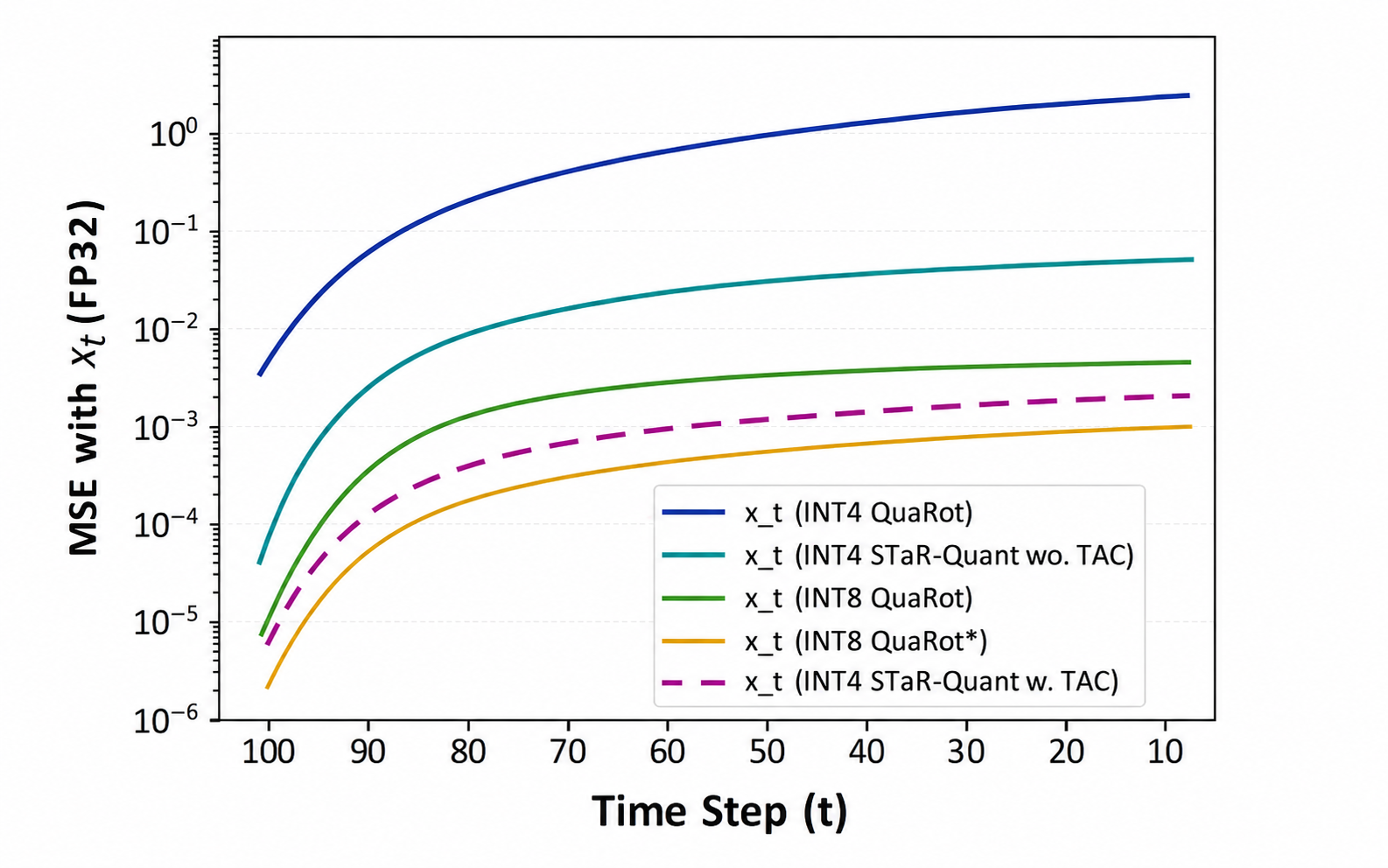}
    \caption{
    Temporal quantization error across denoising steps. 
    We report the MSE between the quantized hidden state and the FP16 hidden state at each denoising step. 
    STaR-Quant with TAC consistently reduces the accumulated error compared with the variant without TAC. 
    QuaRot$^\ast$ denotes the setting where the softmax-value multiplication in attention is kept unquantized.
    }
    \label{fig:temporal_mse}
\end{figure}

After SGAT mitigates state-dependent activation errors in quantized projections, 
DLLMs may still suffer from temporal error accumulation along the iterative denoising trajectory. 
Prior analysis suggests that the softmax-value multiplication in attention is particularly sensitive to low-bit quantization. 
To address this remaining temporal error source, we introduce Temporal Attention Compensation (TAC), which corrects the quantized attention representation before it enters the attention output projection, without modifying the attention computation itself. (More details are provided in Appendix~\ref{app:tac_details}.)

For an attention layer at denoising step $t$, let
\begin{equation}
Z_t^{\mathrm{FP}} =
\mathrm{Softmax}
\left(
\frac{Q_tK_t^\top}{\sqrt{d_h}}
\right)V_t
\end{equation}
denote the full-precision output of the softmax-value multiplication, and let $\hat{Z}_t$ be its quantized counterpart. 
Since DLLM generation proceeds through iterative denoising, the distribution shift in $\hat{Z}_t$ can affect subsequent predictions. 
TAC therefore compensates $\hat{Z}_t$ before the output projection, where the attention representation is directly consumed by the next linear operator. 
As shown in Fig.~\ref{fig:temporal_mse}, this targeted compensation consistently reduces the hidden-state error accumulated across denoising steps.

\paragraph{Block-wise affine compensation.}
A direct way to compensate $\hat{Z}_t$ is to learn a dense affine mapping over the whole hidden dimension. 
However, such a mapping is expensive and unstable to estimate from limited reference activations in post-training quantization. 
A purely channel-wise correction is more stable but too restrictive, since the softmax-value multiplication mixes token information and induces channel correlations. 
To balance correction capacity and estimation stability, TAC adopts a block-wise affine mapping. 
We split the hidden dimension into $B$ blocks with block size $g$ and correct each block independently:
\begin{equation}
Z_{t,b}^{\mathrm{tac}}
=
\hat{Z}_{t,b}M_b^\top+\mathbf{1}d_b^\top,
\quad b=1,\ldots,B .
\end{equation}
Here, $\hat{Z}_{t,b}$ is the $b$-th block of the quantized attention representation, 
$M_b\in\mathbb{R}^{g\times g}$ is the block compensation matrix, and 
$d_b\in\mathbb{R}^{g}$ is the bias term. 
This block-diagonal form preserves local channel interactions within each block while avoiding the cost and unreliable estimation of a dense $d\times d$ mapping.

\paragraph{Closed-form estimation.}
To keep TAC lightweight, we estimate the compensation parameters by matching the first- and second-order statistics between full-precision and quantized attention representations. 
After SGAT is optimized, we collect paired activations $(Z_t^{\mathrm{FP}},\hat{Z}_t)$ at the TAC insertion point. 
For each block, let 
$(\mu_b^{\mathrm{FP}},\Sigma_b^{\mathrm{FP}})$ and 
$(\mu_b^{\mathrm{Q}},\Sigma_b^{\mathrm{Q}})$ denote the mean and covariance statistics of the full-precision and quantized representations, respectively. 
The compensation parameters are computed in closed form:
\begin{equation}
\begin{aligned}
M_b &=
(\Sigma_b^{\mathrm{FP}})^{\frac{1}{2}}
(\Sigma_b^{\mathrm{Q}}+\lambda I)^{-\frac{1}{2}}, \\
d_b &= \mu_b^{\mathrm{FP}} - M_b\mu_b^{\mathrm{Q}} .
\end{aligned}
\end{equation}
The damping term $\lambda I$ stabilizes the inverse square-root of the quantized covariance. 
With this mapping, $M_b$ aligns the covariance structure and $d_b$ restores the mean, making the compensated representation closer to the full-precision attention output before projection.

Since the statistics are estimated from limited reference activations, the closed-form solution may contain noisy directions. 
We therefore further stabilize $M_b$ by smoothing its singular values and shrinking the final matrix toward the identity. 
This prevents TAC from amplifying unreliable directions while retaining its ability to correct temporal distribution shifts. 
Overall, TAC is inserted only before the attention output projection, providing a targeted and lightweight correction for temporal error accumulation in quantized DLLMs.

\begin{table*}[!t]
\centering
\caption{Results of RTN, AWQ, QuaRot, DLLMQuant and our STaR-Quant with 4-bit weight and activation quantization among 9 tasks
on LLADA-8B, LLADA-1.5-8B, DREAM-7B. DLLMQuant$^{+}$ denotes DLLMQuant based on AWQ, and DLLMQuant$^{++}$ denotes DLLMQuant based on QuaRot.}
\label{tab:quant_results}
\scriptsize
\setlength{\tabcolsep}{4pt}
\renewcommand{\arraystretch}{1.12}

\begin{threeparttable}
\resizebox{\textwidth}{!}{
\begin{tabular}{lcccccccccc}
\toprule
\textbf{Method}
& \multicolumn{9}{c}{\textbf{Evaluation Benchmarks}}
& \textbf{Avg.} \\
\cmidrule(lr){2-10}
& \textbf{Truth.} & \textbf{Arc.} & \textbf{Hella.} & \textbf{Wino.}
& \textbf{PIQA} & \textbf{MMLU} & \textbf{C-EVAL}
& \textbf{Hum.} & \textbf{GSM8K} & \\
\midrule

\rowcolor{modelrow}
\multicolumn{11}{l}{\textit{\textbf{LLADA-8B}}} \\
FP                  & 47.49 & 44.03 & 54.06 & 74.90 & 74.65 & 65.85 & 69.54 & 32.92 & 67.48 & 58.99 \\
RTN                 & 40.45 & 41.83 & 45.40 & 64.72 & 67.95 & 49.26 & 57.95 & 14.02 & 16.56 & 44.23 \\
AWQ                 & 40.87 & 42.92 & 46.14 & 66.88 & 69.43 & 51.22 & 58.43 & 20.10 & 36.88 & 48.09 \\
DLLMQuant$^{+}$     & 41.53 & 43.44 & 46.51 & 67.87 & 70.12 & 51.72 & 59.38 & 22.13 & 40.66 & 49.26 \\
QuaRot              & 42.53 & 44.20 & 49.76 & 69.85 & 70.75 & 55.96 & 56.32 & 25.33 & 44.57 & 51.03 \\
DLLMQuant$^{++}$    & 43.53 & 44.18 & 51.00 & 71.85 & \textbf{73.94} & 57.77 & 61.22 & 28.92 & 56.25 & 54.29 \\
\rowcolor{oursrow}
\textbf{STaR-Quant} & \textbf{49.12} & \textbf{44.23} & \textbf{52.75} & \textbf{72.92} & 73.85 & \textbf{62.95} & \textbf{64.56} & \textbf{35.98} & \textbf{57.29} & \textbf{57.07} \\

\midrule

\rowcolor{modelrow}
\multicolumn{11}{l}{\textit{\textbf{LLADA-1.5-8B}}} \\
FP                  & 47.20 & 88.50 & 74.70 & 74.80 & 74.86 & 66.00 & 70.05 & 49.40 & 83.30 & 69.86 \\
RTN                 & 39.51 & 81.77 & 56.82 & 65.27 & 66.91 & 48.88 & 58.96 & 23.44 & 36.56 & 53.12 \\
AWQ                 & 40.96 & 82.22 & 66.84 & 67.93 & 69.32 & 51.12 & 60.03 & 30.07 & 57.95 & 58.49 \\
DLLMQuant$^{+}$     & 42.14 & 83.38 & 70.09 & 68.69 & 70.22 & 51.63 & 61.28 & 32.44 & 59.55 & 59.94 \\
QuaRot              & 43.21 & 84.23 & 65.34 & 69.55 & 70.17 & 56.23 & 57.66 & 37.33 & 65.86 & 61.06 \\
DLLMQuant$^{++}$    & 43.87 & 84.18 & 69.20 & 71.74 & \textbf{73.58} & 57.27 & 60.04 & 44.58 & 74.30 & 64.31 \\
\rowcolor{oursrow}
\textbf{STaR-Quant} & \textbf{48.77} & \textbf{85.12} & \textbf{72.43} & \textbf{72.09} & 73.36 & \textbf{63.29} & \textbf{64.34} & \textbf{46.53} & \textbf{76.45} & \textbf{66.93} \\

\midrule

\rowcolor{modelrow}
\multicolumn{11}{l}{\textit{\textbf{DREAM-7B}}} \\
FP                  & 49.76 & 59.80 & 73.30 & 74.50 & 75.66 & 69.50 & 64.89 & 57.90 & 77.20 & 66.94 \\
RTN                 & 41.25 & 54.34 & 58.32 & 64.52 & 66.26 & 51.51 & 49.89 & 28.90 & 30.82 & 49.53 \\
AWQ                 & 43.66 & 57.82 & 65.57 & 67.13 & 68.96 & 55.50 & 53.27 & 33.14 & 48.98 & 54.89 \\
DLLMQuant$^{+}$     & 44.14 & 58.38 & 66.93 & 68.66 & 69.53 & 57.63 & 54.21 & 35.12 & 51.14 & 56.19 \\
QuaRot              & 47.58 & 58.18 & 67.13 & 70.05 & 70.36 & \textbf{69.50} & 53.19 & 34.48 & 59.20 & 58.85 \\
DLLMQuant$^{++}$    & 47.86 & 59.43 & 70.14 & 71.54 & 72.09 & \textbf{69.50} & 55.89 & 44.50 & 66.17 & 61.90 \\
\rowcolor{oursrow}
\textbf{STaR-Quant} & \textbf{48.72} & \textbf{59.76} & \textbf{71.32} & \textbf{72.98} & \textbf{74.34} & 69.37 & \textbf{58.79} & \textbf{47.43} & \textbf{69.57} & \textbf{63.59} \\

\bottomrule
\end{tabular}
}
\end{threeparttable}
\end{table*}

\section{Experiments}

\subsection{Experimental settings}
All experiments are conducted on NVIDIA A40 GPUs, unless otherwise specified. As
STaR-Quant is an efficient post-training quantization (PTQ) framework, it eliminates the need for any fine-tuning.

\textbf{Evaluated DLLMs and Quantization Baselines.} We conduct comprehensive evaluations on three recent diffusion-based language models, LLaDA-8B~\cite{nie2025large}, LLaDA-1.5-8B~\cite{zhu2025llada15variancereducedpreference} and Dream 7B~\cite{ye2025dream}. Our main baselines include vanilla RTN and three PTQ approaches for LLMs: AWQ~\cite{lin2024awq}, QuaRot~\cite{ashkboos2024quarot} and DLLMQuant~\cite{xu2025dllmquant}. For calibration, we use 128 segments sampled from the WinoGrande dataset. In the case of QuaRot, we follow the implementation strategy of its official repository and apply GPTQ, a Hessian-based weight compensation method, to adjust the quantized weights.

\textbf{Evaluation Benchmarks.} We evaluate the performance of STaR-Quant across three task categories, following the evaluation setup of the LLaDA paper: 1) General knowledge and reasoning tasks, including TruthfulQA-MC2~\cite{lin2022truthfulqa}, ARC-Challenge~\cite{clark2018arc}, HellaSwag~\cite{zellers2019hellaswag}, WinoGrande~\cite{sakaguchi2019winogrande}, PIQA~\cite{bisk2019piqa}, MMLU~\cite{hendrycks2021mmlu}, and C-EVAL~\cite{huang2023ceval}, where accuracy is used as the evaluation metric; 2) Mathematical reasoning tasks, represented by GSM8K~\cite{cobbe2021gms8k}, which assesses multistep reasoning ability; and 3) Code generation tasks, represented by HumanEval~\cite{chen2021humaneval}, which evaluates code generation capability. Together, these benchmarks provide a comprehensive assessment of our method from multiple perspectives.
\subsection{Main Results}
The results in Table~\ref{tab:quant_results} lead to three observations.
First, STaR-Quant consistently outperforms existing PTQ methods.
Under W4A4 quantization, STaR-Quant achieves the best average performance on all three DLLMs. 
It improves over DLLMQuant$^{++}$ by 2.78, 2.62, and 1.69 average points on LLaDA-8B, LLaDA-1.5-8B, and Dream-7B, respectively.
Compared with RTN, AWQ, and QuaRot, the improvement is more significant, showing that directly applying conventional LLM quantization methods is insufficient for dLLMs.
Second, STaR-Quant better preserves general knowledge capability.
On benchmarks such as TruthfulQA, ARC, HellaSwag, WinoGrande, PIQA, MMLU, and C-EVAL, STaR-Quant achieves strong performance across different models. 
In particular, the improvements on MMLU and C-EVAL indicate that our method better maintains multi-domain knowledge and language understanding under low-bit quantization.
Third, STaR-Quant is more robust on reasoning and generation tasks.
GSM8K and HumanEval are more sensitive to quantization errors because they require multi-step reasoning and precise code generation. 
STaR-Quant consistently improves these tasks over DLLMQuant$^{++}$. 
For example, on LLaDA-8B, HumanEval increases from 28.92 to 35.98 and GSM8K increases from 56.25 to 57.29. 
On Dream-7B, HumanEval improves from 44.50 to 47.43 and GSM8K improves from 66.17 to 69.57. 
These gains suggest that state-guided activation transformation and temporal attention compensation help preserve the intermediate representations required by complex generation tasks. Additional W8A8 quantization results are provided in Appendix~\ref{app:results}.

\subsection{Ablation Study}

Table~\ref{tab:ablation} reports the contribution of each component in STaR-Quant. 
The full method achieves the best quantized average performance of 57.07, demonstrating the effectiveness of combining state-aware transformation and temporal compensation. 
When TAC is removed, the average score drops to 55.43, indicating that compensating the quantized attention representation before the output projection is important for suppressing temporal error accumulation. 
When SGAT is removed, the score decreases to 56.39, showing that State-Guided Activation Transformation helps reduce the activation mismatch between masked and unmasked tokens. 
These results suggest that the two modules are complementary: SGAT improves state-wise activation consistency, while TAC improves temporal attention consistency.
\begin{table}[H]
\centering
\caption{Component ablation of STaR-Quant under W4A4 quantization on LLaDA-8B. Avg. denotes the average accuracy over nine tasks. $\Delta\downarrow$ denotes the accuracy drop compared with FP.}
\label{tab:ablation}
\begin{tabular*}{\columnwidth}{@{\extracolsep{\fill}}lcc}
\toprule
\textbf{Method} & \textbf{Avg.} & \textbf{$\Delta\downarrow$} \\
\midrule
FP & 58.99 & 0.00 \\
STaR-Quant w/o TAC & 55.43 & 3.56 \\
STaR-Quant w/o SGAT & 56.39 & 2.60 \\
STaR-Quant & \textbf{57.07} & \textbf{1.92} \\
\bottomrule
\end{tabular*}
\end{table}

We further study the effect of the TAC block size in Table~\ref{tab:tac_block}. 
The block size controls the trade-off between the expressive capacity and the estimation stability of the block-wise affine compensation. 
A small block size, such as $g=4$ or $g=8$, provides limited cross-channel modeling ability and leads to clearly lower average performance. 
Increasing the block size improves the result, and the best performance is achieved at $g=16$, reaching an average accuracy of 57.07. 
However, further increasing the block size to $g=32$ or $g=64$ slightly degrades the performance, which suggests that overly large blocks may introduce less stable covariance estimation under limited calibration data. 
Therefore, we use $g=16$ as the default TAC block size, as it provides the best balance between compensation capacity and calibration robustness.

\begin{table}[H]
\centering
\small
\caption{Effect of TAC block size.}
\label{tab:tac_block}
\begin{tabular}{lccccc}
\toprule
Block size $g$ & 4 & 8 & 16 & 32 & 64 \\
\midrule
Avg. & 52.48 & 54.66 & 57.07 & 56.39 & 56.03 \\
\bottomrule
\end{tabular}
\end{table}

\subsection{Memory and Speedup}
\textbf{Inference Speedup.}
To evaluate the practical efficiency of STaR-Quant, we follow the measurement protocol and W4A4 kernel implementation used in ~\cite{ashkboos2024quarot}. 
As shown in Table~\ref{tab:speed_memory}, STaR-Quant consistently accelerates inference across different DLLMs. 
Specifically, it achieves 1.65$\times$, 1.64$\times$, and 1.69$\times$ speedup on LLaDA, LLaDA-1.5, and Dream, respectively. 
The average speedup reaches 1.66$\times$, showing that the proposed state-guided transformation and TAC do not introduce prohibitive runtime overhead under W4A4 execution.

\textbf{Memory Consumption.}
We further measure the peak GPU memory usage during inference. 
As reported in Table~\ref{tab:speed_memory}, STaR-Quant significantly reduces memory consumption for all evaluated DLLMs. 
For LLaDA and LLaDA-1.5, the memory usage decreases from about 15.9GB to 5.2GB, corresponding to 3.05$\times$ memory saving. 
For Dream, the memory usage is reduced from 13.95GB to 4.44GB, achieving 3.14$\times$ memory saving. 
These results demonstrate that STaR-Quant effectively lowers both computational and memory costs, making low-bit DLLM deployment more practical while maintaining strong accuracy retention.
\begin{table}[t]
\centering
\caption{Speedup and memory saving of three DLLMs, compared between 4-bit implementation and FP16.}
\label{tab:speed_memory}
\renewcommand{\arraystretch}{1.30}
\resizebox{\columnwidth}{!}{
\begin{tabular}{lcccccc}
\toprule
\multirow{2}{*}{\textbf{MODEL}} 
& \multicolumn{3}{c}{\textbf{Speed (Tokens/s)}} 
& \multicolumn{3}{c}{\textbf{Memory (GB)}} \\
\cmidrule(lr){2-4} \cmidrule(lr){5-7}
& \textbf{FP} 
& \textbf{Quant} 
& \textbf{SpeedUp} 
& \textbf{FP} 
& \textbf{Quant} 
& \textbf{Mem. Sav.} \\
\midrule
LLADA     & 34.59 & 57.07 & \textbf{1.65} & 15.89 & 5.20 & \textbf{3.05} \\
LLADA-1.5 & 35.55 & 58.30 & \textbf{1.64} & 15.88 & 5.20 & \textbf{3.05} \\
DREAM     & 23.27 & 39.33 & \textbf{1.69} & 13.95 & 4.44 & \textbf{3.14} \\
\bottomrule
\end{tabular}
}
\end{table}

\section{Conclusion}

We present \textbf{STaR-Quant}, a state-time consistent PTQ framework for low-bit weight-activation quantization of DLLMs. 
It targets two DLLM-specific challenges: state-dependent activation disparity between masked and unmasked tokens, and temporal error accumulation during iterative denoising. 
To address them, SGAT maps different token states into state-aware activation transformation spaces while sharing a unified transformed weight, and TAC compensates the quantized attention representation before the output projection to reduce temporal quantization error. 
Experiments on LLaDA, LLaDA-1.5, and Dream show that STaR-Quant preserves accuracy better than strong PTQ baselines, with clear gains on knowledge, reasoning, and code generation benchmarks. 
Ablations and temporal error analysis validate both components, demonstrating state-time consistency as an effective principle for efficient DLLM quantization.

\section*{Limitations}

STaR-Quant mainly focuses on W4A4 post-training quantization for representative masked-denoising DLLMs. 
Although experiments on LLaDA, LLaDA-1.5, and Dream demonstrate consistent improvements, further evaluation on more diffusion architectures, model scales, and denoising schedules is needed to fully validate its generality. 
Moreover, more aggressive settings such as 3-bit or 2-bit weight-activation quantization remain challenging and may require stronger state-aware transformation or temporal compensation strategies.
Another limitation lies in system-level deployment. 
TAC introduces a lightweight block-wise affine compensation before the attention output projection, and SGAT applies state-conditioned activation transformation. 
While both components are efficient in principle, their practical speedup can be further improved with dedicated fused kernels. 
Finally, our current state modeling is based on predefined masked and unmasked tokens. 
Extending STaR-Quant to finer-grained token states, confidence-aware states, or multimodal diffusion models is an interesting direction for future work.

% Bibliography entries for the entire Anthology, followed by custom entries
%\bibliography{anthology,custom}
% Custom bibliography entries only
\bibliography{custom}

\appendix

\section{Computation Consistency of SGAT}
\label{app:sgat_consistency}

In this section, we show that State-Guided Activation Transformation (SGAT) can use state-conditioned activation transformations while keeping a unified weight-side transformation. 
For clarity, we omit the layer index and consider a linear layer with input activation $X\in\mathbb{R}^{L\times d}$ and weight $W\in\mathbb{R}^{C\times d}$. 
Let $X^m$ and $X^u$ denote the activations of masked and unmasked tokens, respectively.

SGAT constructs a full transformation basis
\begin{equation}
P=[U_{\mathrm{sh}}, U_m, U_u]\in\mathbb{R}^{d\times d},
\end{equation}
where $U_{\mathrm{sh}}$, $U_m$, and $U_u$ correspond to the shared, masked-specific, and unmasked-specific subspaces. 
We assume that these bases form an orthonormal decomposition, i.e.,
\begin{equation}
P^\top P = PP^\top = I .
\end{equation}
For a general invertible basis, $P^\top$ can be replaced by $P^{-1}$ without changing the following analysis.

The state-conditioned gates are defined as
\begin{equation}
\begin{aligned}
G_m &= \mathrm{diag}(\mathbf{1}_{d_{\mathrm{sh}}},
                     \mathbf{1}_{d_m},
                     \mathbf{0}_{d_u}), \\
G_u &= \mathrm{diag}(\mathbf{1}_{d_{\mathrm{sh}}},
                     \mathbf{0}_{d_m},
                     \mathbf{1}_{d_u}).
\end{aligned}
\end{equation}
Thus, masked tokens use the shared and masked-specific subspaces, while unmasked tokens use the shared and unmasked-specific subspaces. 
For state $s\in\{m,u\}$, the activation-side transformation is
\begin{equation}
\Phi_s(X^s)=X^sPG_s .
\end{equation}
The weight side, however, uses only one unified transformation:
\begin{equation}
\widetilde{W}=P^\top W^\top .
\end{equation}

We now show when this state-conditioned computation preserves the original full-precision linear projection. 
Define the inactive gate of state $s$ as
\begin{equation}
G_s^\perp = I-G_s .
\end{equation}
If the activation of state $s$ has no component in its inactive subspace after transformation, i.e.,
\begin{equation}
X^sPG_s^\perp = 0 ,
\label{eq:inactive_zero}
\end{equation}
then
\begin{equation}
X^sP = X^sP(G_s+G_s^\perp)=X^sPG_s .
\end{equation}
Using $PP^\top=I$, the original linear computation can be rewritten as
\begin{equation}
\begin{aligned}
X^sW^\top
&= X^sPP^\top W^\top \\
&= X^sP(G_s+G_s^\perp)P^\top W^\top \\
&= X^sPG_sP^\top W^\top \\
&= \Phi_s(X^s)\widetilde{W}.
\end{aligned}
\end{equation}
Therefore, under the ideal subspace allocation condition in Eq.~\ref{eq:inactive_zero}, SGAT preserves the original linear computation while using different activation-side transformations for different token states and a single shared weight-side transformation.

In practice, Eq.~\ref{eq:inactive_zero} may not hold exactly. 
The deviation introduced by suppressing the inactive subspace can be written as
\begin{equation}
\Delta_s
=
X^sW^\top-\Phi_s(X^s)\widetilde{W}
=
X^sPG_s^\perp P^\top W^\top .
\end{equation}
This term measures the information leakage into the subspace assigned to the other token state. 
SGAT reduces this deviation by optimizing the transformation basis with the layer-wise reconstruction objective:
\begin{equation}
\begin{aligned}
\mathcal{L}_{\mathrm{state}}
&=
\mathbb{E}_{t}
\sum_{s\in\{m,u\}}
\left\|
Y_{t}^{s,\mathrm{FP}} - \hat{Y}_{t}^{s}
\right\|_{F}^{2}, \\
\hat{Y}_{t}^{s}
&=
Q_a\!\left(X_{t}^{s} P G_s\right)
Q_w\!\left(P^\top W^\top\right).
\end{aligned}
\end{equation}
This objective jointly accounts for the subspace allocation error and the quantization error.
As a result, the learned transformation encourages masked and unmasked activations to occupy their corresponding subspaces while sharing the same transformed weight.

Importantly, the weight-side transform $P^\top W^\top$ is independent of the token state. 
Thus, SGAT does not require maintaining separate transformed weights for masked and unmasked tokens. 
After optimization, the transformed weight can be precomputed and quantized once, while state-conditioned gates are applied only on the activation side during inference.
\section{Implementation Details}
\label{app:details}
For each quantized linear layer, SGAT first partitions the input
tokens into masked and unmasked groups according to their current
denoising states. Let $d$ denote the hidden dimension. We construct
an orthogonal transformation basis
$P=[U_{sh}, U_m, U_u]$, where $U_{sh}$ is shared by both token
states, while $U_m$ and $U_u$ are reserved for masked and unmasked
tokens, respectively. Following the subspace allocation strategy of
FreeAct, we set the state-specific dimensions to
$d_m=d_u=d/32$ by default and assign the remaining dimensions to
the shared subspace, i.e.,
$d_{sh}=d-d_m-d_u$. This allocation keeps most transformation
capacity in the shared subspace, which captures common activation
components across token states, while reserving a small number of
dimensions for state-specific outlier patterns. In this way, SGAT
can adapt to the distributional mismatch between masked and
unmasked tokens without duplicating the transformed weights.

For initialization, we sample a random Gaussian matrix and
orthogonalize it to obtain the initial transformation basis $P$.
During calibration, $P$ is constrained to remain orthogonal, so that
the weight-side transformation can be represented by $P^\top$ and
folded into the linear weight. Given the state gates
$G_m=\mathrm{diag}(\mathbf{1}_{d_{sh}},\mathbf{1}_{d_m},
\mathbf{0}_{d_u})$ and
$G_u=\mathrm{diag}(\mathbf{1}_{d_{sh}},\mathbf{0}_{d_m},
\mathbf{1}_{d_u})$, masked tokens use $P_m=PG_m$ and unmasked
tokens use $P_u=PG_u$. The zero-padded inactive subspaces prevent
one token state from using the state-specific dimensions assigned
to the other state.

We optimize SGAT in a layer-wise post-training calibration manner.
The original model parameters are frozen, and only the
transformation basis is updated. For each layer, we collect paired
full-precision and quantized activations from the calibration set.
The full-precision output is computed by the original linear layer,
while the quantized output is computed using the state-conditioned
activation transformation and the shared transformed quantized
weight. We minimize the reconstruction objective in Eq. (15) using
AdamW with learning rate $1\times10^{-3}$, batch size 4, and 15
calibration epochs. This objective jointly encourages masked and
unmasked activations to occupy their corresponding subspaces and
reduces the quantization error under the unified weight-side
transformation.

After calibration, the weight-side transformation is precomputed
and folded into the quantized weight as $W_f=Q_w(P^\top W^\top)$.
Therefore, SGAT does not maintain separate weights for masked and
unmasked tokens during inference. At runtime, we only apply the
state-conditioned activation gates according to the token mask
states, quantize the transformed activations, multiply them with the
shared quantized weight, and scatter the outputs back to the
original token order.
\section{Detailed Implementation of Temporal Attention Compensation}
\label{app:tac_details}

In this section, we provide the implementation details of Temporal
Attention Compensation (TAC). TAC is inserted after the
softmax-value multiplication in self-attention and before the
attention output projection. Therefore, for an attention layer, the
compensated representation is computed as
\[
    \mathrm{Softmax}(QK^\top / \sqrt{d_h})V
    \;\rightarrow\;
    \mathrm{TAC}
    \;\rightarrow\;
    W_o,
\]
where \(W_o\) denotes the attention output projection. This insertion
point allows TAC to correct the quantized attention representation
directly before it is consumed by the following linear projection.

\paragraph{Block-wise statistics.}
Let \(\hat Z_t \in \mathbb{R}^{N \times C}\) denote the quantized
attention representation collected at the TAC insertion point, where
\(N\) is the total number of calibration tokens after flattening the
batch and sequence dimensions, and \(C\) is the hidden dimension. To
avoid estimating a full \(C \times C\) covariance matrix, we split the
hidden dimension into \(B\) non-overlapping blocks with block size
\(g\), where \(C = B g\). By default, we set \(g=16\). If \(C\) is not
divisible by the default block size, we choose the largest block size
not exceeding 16 that divides \(C\).

For each layer and each block \(b\), we collect paired full-precision
and quantized attention representations,
\[
    Z^{\mathrm{FP}}_{b}, \quad \hat Z^{\mathrm{Q}}_{b}
    \in \mathbb{R}^{N \times g}.
\]
The full-precision pass disables activation and weight quantizers,
while keeping the learned transformation modules enabled. The
quantized pass enables quantization and uses the same calibration
samples. We then estimate the population mean and covariance for each block:
\begin{equation}
\begin{aligned}
\mu^{\mathrm{FP}}_b
&=
\frac{1}{N}
\sum_{i=1}^{N} z^{\mathrm{FP}}_{i,b},
\\
\Sigma^{\mathrm{FP}}_b
&=
\frac{1}{N}
\sum_{i=1}^{N}
z^{\mathrm{FP}}_{i,b}
{z^{\mathrm{FP}}_{i,b}}^\top
-
\mu^{\mathrm{FP}}_b
{\mu^{\mathrm{FP}}_b}^\top .
\end{aligned}
\end{equation}
and similarly,
\[
    \mu^{\mathrm{Q}}_b,
    \qquad
    \Sigma^{\mathrm{Q}}_b
\]
for the quantized representation.

\paragraph{Damped covariance matching.}
TAC estimates a block-wise affine mapping that aligns the first- and
second-order statistics of the quantized representation with its
full-precision counterpart. Since the covariance estimated from a
small calibration set can be ill-conditioned, we add a damping term to
the quantized covariance:
\[
    \lambda_b
    =
    \rho \cdot
    \max \left(
    \frac{\mathrm{Tr}(\Sigma^{\mathrm{Q}}_b)}{g},
    \epsilon
    \right),
\]
where \(\rho\) is the damping coefficient and \(\epsilon\) is a small
constant for numerical stability. We use
\(\rho=3\times 10^{-4}\) and \(\epsilon=10^{-8}\) by default. The
damped quantized covariance is
\[
    \widetilde{\Sigma}^{\mathrm{Q}}_b
    =
    \Sigma^{\mathrm{Q}}_b + \lambda_b I.
\]

The initial compensation matrix is computed by covariance matching:
\[
    M_b
    =
    \left(\Sigma^{\mathrm{FP}}_b\right)^{1/2}
    \left(\widetilde{\Sigma}^{\mathrm{Q}}_b\right)^{-1/2}.
\]
Matrix square roots and inverse square roots are computed through
symmetric eigendecomposition. Specifically, for a positive
semi-definite matrix \(A\), we compute
\[
    A^p
    =
    Q \, \mathrm{diag}
    \left(
    \max(\lambda_i, 10^{-12})^p
    \right) Q^\top,
\]
where \(A = Q \mathrm{diag}(\lambda_i) Q^\top\). The eigenvalue
clamping prevents numerical instability caused by very small
eigenvalues.

\begin{table*}[htbp]
\centering
\caption{Model performance  under 8-bit weight and activation
quantization.}
\label{tab:quant_results2}
\resizebox{\textwidth}{!}{
\begin{tabular}{llcccccccccc}
\toprule
\textbf{Model} & \textbf{Method} 
& \textbf{Truth.} & \textbf{Arc.} & \textbf{Hella.} & \textbf{Wino.} 
& \textbf{PIQA} & \textbf{MMLU} & \textbf{C-EVAL} 
& \textbf{Hum.} & \textbf{GSM8K} & \textbf{Avg.} \\
\midrule

\multirow{2}{*}{LLADA}
& FP  & 47.49 & 44.03 & 54.06 & 74.90  & 74.65 & 65.85 & 69.54 & 32.92 & 67.48 & 58.99 \\
& STaR-Quant & 48.56 & 44.25 & 54.40 & 68.72 & 74.23 & 64.36 & 67.87 & 36.09 & 65.74 & 58.25 \\

\midrule

\multirow{2}{*}{LLADA-1.5}
& FP         & 47.20  & 88.50  & 74.70  & 74.80  & 74.86 & 66.00  & 70.05  & 49.4  & 83.30  & 69.86 \\
& STaR-Quant & 49.72 & 86.50 & 74.31 & 73.98 & 74.24 & 65.79 & 68.96 & 47.55  & 79.37 & 68.94 \\

\midrule

\multirow{2}{*}{DREAM}
& FP         & 49.76 & 59.80 & 73.30 & 74.50 & 75.66 & 69.50  & 64.89 & 57.90  & 77.2  & 66.94 \\
& STaR-Quant & 49.02 & 60.28 & 72.89 & 73.50 & 74.23 & 70.11  & 62.89 & 51.76 & 73.44 & 65.35 \\

\bottomrule
\end{tabular}
}
\end{table*}

\paragraph{Singular value smoothing.}
The closed-form covariance matching solution can overfit noisy
directions in the calibration set. To stabilize the compensation
matrix, we smooth the singular values of \(M_b\). Let
\[
    M_b = U_b \mathrm{diag}(\sigma_{b}) V_b^\top
\]
be the singular value decomposition of \(M_b\), and let
\[
    \bar{\sigma}_b = \frac{1}{g}\sum_{i=1}^{g}\sigma_{b,i}
\]
be the mean singular value in block \(b\). We replace each singular
value by
\[
    \sigma'_{b,i}
    =
    (1-\eta)\sigma_{b,i} + \eta \bar{\sigma}_b,
\]
where \(\eta\) controls the strength of singular value smoothing. We
use \(\eta=0.15\) by default. The smoothed matrix is then reconstructed
as
\[
    M^{\mathrm{sv}}_b
    =
    U_b \mathrm{diag}(\sigma'_b) V_b^\top.
\]
When \(\eta=0\), singular value smoothing is disabled.

\paragraph{Identity shrinkage.}
We further shrink the smoothed compensation matrix toward the identity
matrix:
\[
    M^{\mathrm{final}}_b
    =
    (1-\alpha) M^{\mathrm{sv}}_b + \alpha I.
\]
We set \(\alpha=0.55\) by default. This shrinkage prevents TAC from
over-correcting unreliable covariance directions. When \(\alpha=0\),
TAC fully uses the covariance matching solution; when \(\alpha=1\),
the compensation matrix degenerates to identity and only mean
correction remains.

\paragraph{Bias estimation.}
After obtaining the final block-wise compensation matrix, we compute
the bias term by matching the mean of the compensated quantized
representation to the full-precision representation:
\[
    d_b
    =
    \mu^{\mathrm{FP}}_b
    -
    M^{\mathrm{final}}_b \mu^{\mathrm{Q}}_b.
\]
Therefore, for a token representation \(z_b\) in block \(b\), TAC
applies the following affine correction under the column-vector
notation:
\[
    z^{\mathrm{TAC}}_b
    =
    M^{\mathrm{final}}_b z_b + d_b.
\]
In implementation, activations are stored in row-major format, and the
equivalent computation is
\[
    Z^{\mathrm{TAC}}_b
    =
    \hat Z^{\mathrm{Q}}_b
    {M^{\mathrm{final}}_b}^{\top}
    +
    \mathbf{1}{d_b}^{\top}.
\]

\paragraph{Default hyperparameters.}
Unless otherwise specified, TAC uses the following hyperparameters:
block size \(g=16\), damping coefficient
\(\rho=3\times 10^{-4}\), singular value smoothing coefficient
\(\eta=0.15\), identity shrinkage coefficient \(\alpha=0.55\), and at
most 64 calibration samples for estimating the statistics. TAC is
estimated after SGAT calibration, and its parameters are fixed during
inference.
\section{Additional Results}
Table~\ref{tab:quant_results2} reports additional results under W8A8 quantization. 
Compared with the FP models, STaR-Quant preserves most of the original performance across all three dLLMs. 
Specifically, the average scores only drop from 58.99 to 58.25 on LLaDA, from 69.86 to 68.94 on LLaDA-1.5, and from 66.94 to 65.35 on Dream. 
The small average degradation indicates that STaR-Quant remains stable under a less aggressive quantization setting. 
Moreover, STaR-Quant even slightly improves over FP on several benchmarks, such as TruthfulQA and ARC on LLaDA, TruthfulQA on LLaDA-1.5, and ARC and MMLU on Dream. 
These results suggest that the proposed state-time consistent quantization framework can effectively retain model capability under W8A8 quantization, with only minor accuracy degradation on average.
\label{app:results}

\section{Additional Visualizations}
\begin{figure*}[t]
    \centering
    \includegraphics[width=\textwidth]{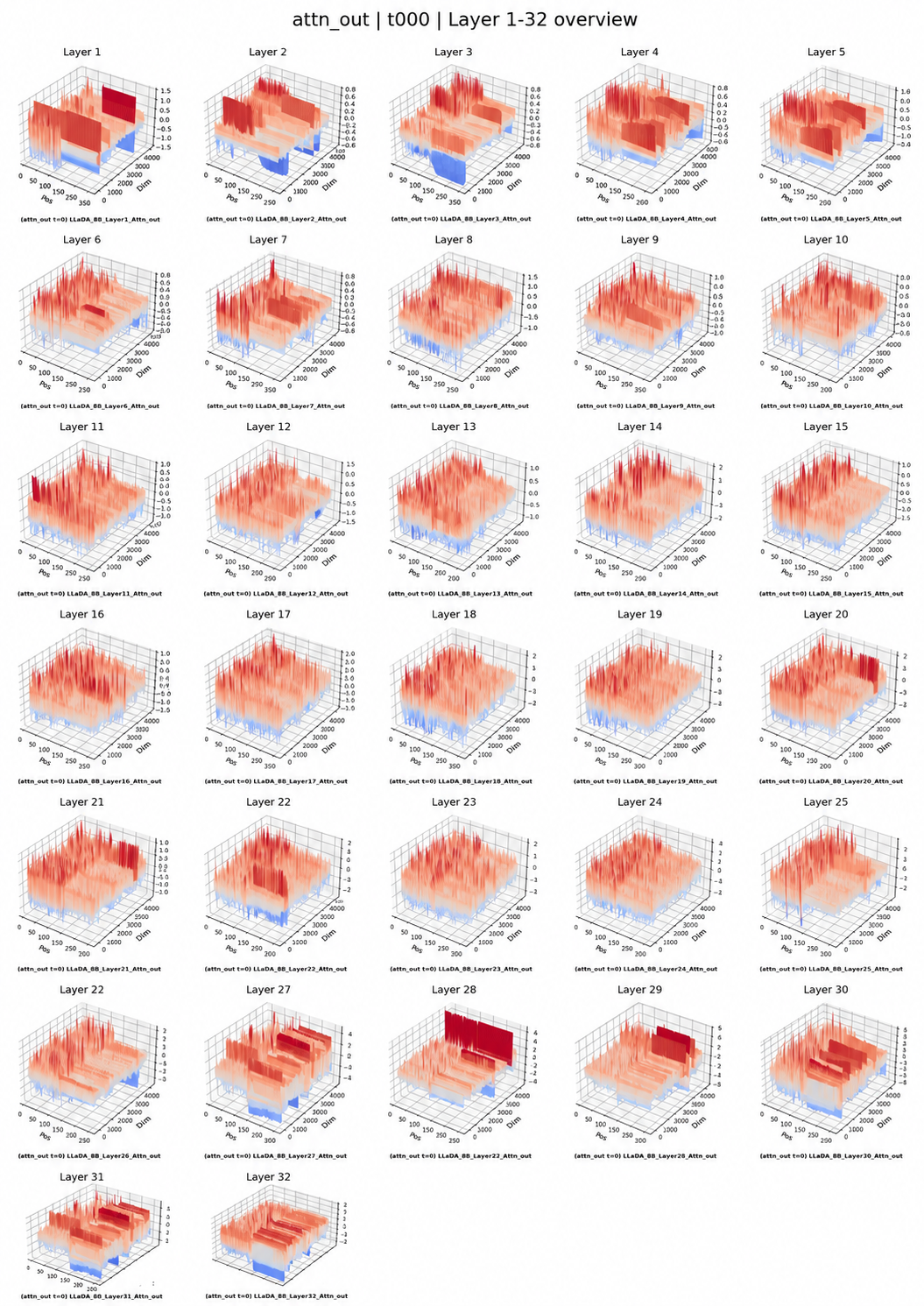}
    \caption{Layer-wise overview of attn\_out at $t=0$ for LLaDA-8B.}
    \label{fig:attn-out-overview}
\end{figure*}

Fig.~\ref{fig:attn-out-overview} visualizes the layer-wise activation
distribution of the attention output input (\texttt{attn\_out}) at
denoising step $t=0$ for LLaDA-8B before applying SGAT. 
The untransformed activations show clear layer-dependent variations
and pronounced outlier patterns, indicating that the attention
representations are difficult to quantize directly under low-bit
weight-activation quantization. 
These observations further support the need for state-guided
activation transformation, which maps masked and unmasked token
activations into more suitable quantization spaces before the
quantized linear projection.

\end{document}